\documentclass[journal]{IEEEtran}
\usepackage[utf8]{inputenc}
\usepackage{cite}
\usepackage{hyperref}
\usepackage{multirow}
\usepackage{algorithmic}
\usepackage{orcidlink}
\usepackage{subfigure}
\usepackage{amsmath} 
\usepackage{amsfonts,amssymb}  
\usepackage{colortbl}
\graphicspath{ {./figures/} }
\usepackage{booktabs}
\usepackage{titlesec} 
\titlespacing*{\section}{0pt}{0.5em}{0.5em}
\titlespacing*{\subsection}{0pt}{0.4em}{0.4em}
\usepackage{stfloats}
\begin{document}

\title{Multitask GLocal OBIA-Mamba for Sentinel-2 Landcover Mapping }

\author{Zack Dewis, Yimin Zhu, Zhengsen Xu, Mabel Heffring, Saeid Taleghanidoozdoozan, Kaylee Xiao, Motasem Alkayid, Lincoln Linlin Xu, ~\IEEEmembership{Member,~IEEE} 


\thanks{This work was supported by the Natural
Sciences and Engineering Research Council of Canada (NSERC) under Grant RGPIN-2019-06744.}
\thanks{Lincoln Linlin Xu, Yimin Zhu, Zack Dewis, Zhengsen Xu, Mabel Heffring, Kaylee Xiao, Saeid Taleghanidoozdoozan, are all with the Department of Geomatics Engineering, University of Calgary, Canada (email: (lincoln.xu, yimin.zhu, zachary.dewis, zhengsen.xu, mabel.heffring1)@ucalgary.ca, staleghanidoozdoozan@uwaterloo.ca) (Corresponding author: Lincoln Linlin Xu)}.

\thanks{Motasem Alkayid is also with the Department of Geography, Faculty of Arts, The University of Jordan, Amman, Jordan (email: motasem.alkayid@ucalgary.ca)
}}

\markboth{Journal of \LaTeX\ Class Files,~Vol.~13, No.~9, September~2014}
{Shell \MakeLowercase{\textit{et al.}}: }
\maketitle
\begin{abstract}
    Although Sentinel-2 based land use and land cover (LULC) classification is critical for various environmental monitoring applications, it is a very difficult task due to some key data challenges (e.g., spatial heterogeneity, context information, signature ambiguity). This paper presents a novel Multitask Glocal OBIA-Mamba (MSOM) for enhanced Sentinel-2 classification with the following contributions. First, an object-based image analysis (OBIA) Mamba model (OBIA-Mamba) is designed to reduce redundant computation without compromising fine-grained details by using superpixels as Mamba tokens. Second, a global-local (GLocal) dual-branch convolutional neural network (CNN)-mamba architecture is designed to jointly model local spatial detail and global contextual information. Third, a multitask optimization framework is designed to employ dual loss functions to balance local precision with global consistency. The proposed approach is tested on Sentinel-2 imagery in Alberta, Canada, in comparison with several advanced classification approaches, and the results demonstrate that the proposed approach achieves higher classification accuracy and finer details that the other state-of-the-art methods.
\end{abstract}
\begin{IEEEkeywords}
Deep learning, land use and land cover, OBIA, remote sensing, Sentinel-2, semantic segmentation, superpixels.
\end{IEEEkeywords}

\begin{figure}[h]
    \centering
    \includegraphics[width=0.8\linewidth]{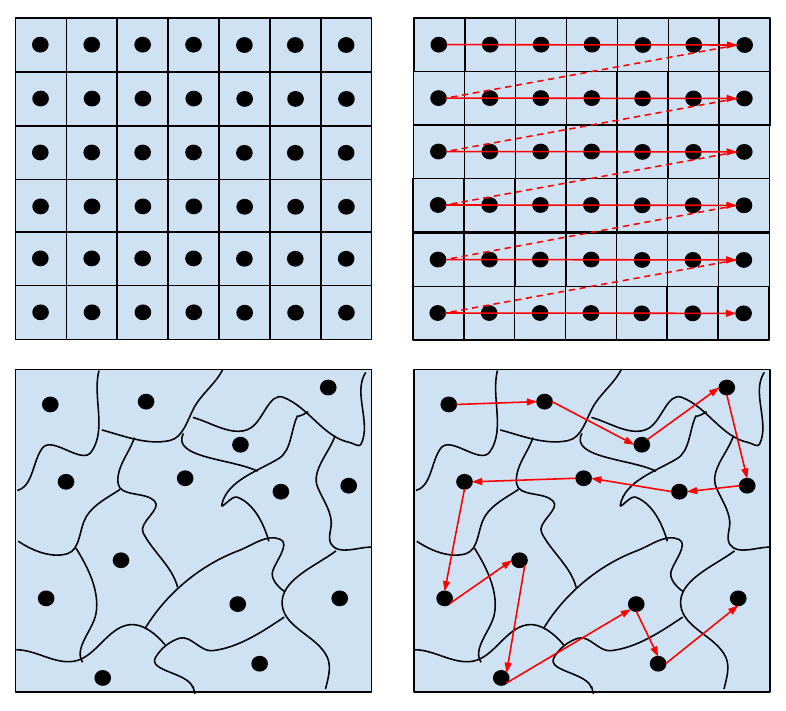}
    \caption{\textbf{Traditional Mamba approaches} (Top) treat each pixel as a token, and scan the tokens in a fixed, predefined, dense and rigid manner, whereas our \textbf{OBIA-Mamba approach} (bottom) treat superpixels/objects as tokens, and builds token sequence in a dynamic, learnable, sparse and adaptable manner, leading to reduced computational cost, improved edge preservation and enhanced longer-range, larger-scale modelling capabilities.  }
    \label{scanning}
\end{figure}
\IEEEpeerreviewmaketitle
\section{Introduction}

The classification of satellite images into accurate pixel-level land use and land cover (LULC) maps is essential for various key applications, such as biodiversity monitoring, urban planning, and environmental management. Nevertheless, accurate and efficient extraction of precise LULC information is challenging due to some key sentinel-2 data characteristics (e.g., spatial heterogeneity, context information, signature ambiguity). Addressing these challenges requires advanced machine learning (ML) and deep learning (DL) models that can accurately learn the most discriminative features for improving classification performances \cite{Wang2020}

Traditional feature extraction methods like Random Forest (RF) and XGBoost have been widely used in LULC classification, relying on engineered features and statistical correlations derived from spectral profiles \cite{Ullah2021}. However, such approaches are fundamentally knowledge-driven and often lack the flexibility for adaptive feature learning, limiting their performance in diverse geographical contexts \cite{Chang2014}. This limitation has prompted a shift towards more sophisticated feature learning frameworks utilizing ML \& DL techniques better suited to handle the intricate and multi-dimensional nature of Sentinel-2 remote sensing data.

Recent advances in deep learning, particularly through the adoption of Convolutional Neural Networks (CNNs), have significantly enhanced the feature extraction capabilities of LULC characteristics. CNNs can effectively capture local spatial patterns due to their hierarchical structure and shared weights for convolutional operations. Fully convolutional networks, such as U-Net, have demonstrated accurate results, especially in complex classification tasks \cite{Shi2024}. However, the inherent locality bias of CNNs can limit their ability to capture longer-range spatial dependencies, essential for accurate LULC mapping \cite{Lam2023}.

To address the challenges posed by local context limitations, transformer architectures have emerged as a viable alternative. They take advantage of self-attention mechanisms that allow models to focus on global features across entire images, which can significantly enhance the learning of large-scale spatial relationships within LULC classification tasks \cite{Fan2022}. However, transformer models have large computational cost, especially in large-scale applications with long input sequence \cite{Sari2017}.

The Mamba models, as a state space approach, are gaining popularity for their ability to maintain high computational efficiency while effectively modeling long-range dependencies, which is crucial in  remote sensing analysis \cite{gu2023mamba}. Nonetheless, optimally constructing token sequences that balance efficiency with the completeness of semantic information remains a significant challenge for Mamba models. Traditional Mamba methods often utilize predetermined scanning systems, which leads to inadequate representation of spatial structures \cite{Stepchenko2017}. Although pixel-level Mamba can better preserve local details, it requires large computational cost. In contrast, patch-based Mamba methods, although have less computational costs, tend to erase critical local detail information (e.g., edges, boundaries and small features) \cite{duan2021novel}. Given the drawbacks of pixel Mamba and patch Mamba, how to use superpixel-based Mamba to strike a good balance between computational efficiency and detail preservation is a critical research topic \cite{yu2022edtrs}.

The integration of superpixels with Mamba can also revolutionize the traditional object-based image analysis (OBIA) techniques, which are widely used in remote sensing image analysis \cite{Ullah2021}. Although traditional OBIA can better capture local structure information, they are inefficient in terms of (1) learning the features to represent the objects/superpixels, and (2) use global attention models (e.g., transformers and mamba) to model the global correlation among the objects/superpixels \cite{zheng2023semantic}. The integration of OBIA with Mamba models in a CNN-Mamba GLocal framework can better address these two challenges and thereby greatly improve the traditional OBIA approach using modern deep learning architectures. 
\begin{figure*}[htbp]
    \centering
    \includegraphics[scale=0.55]{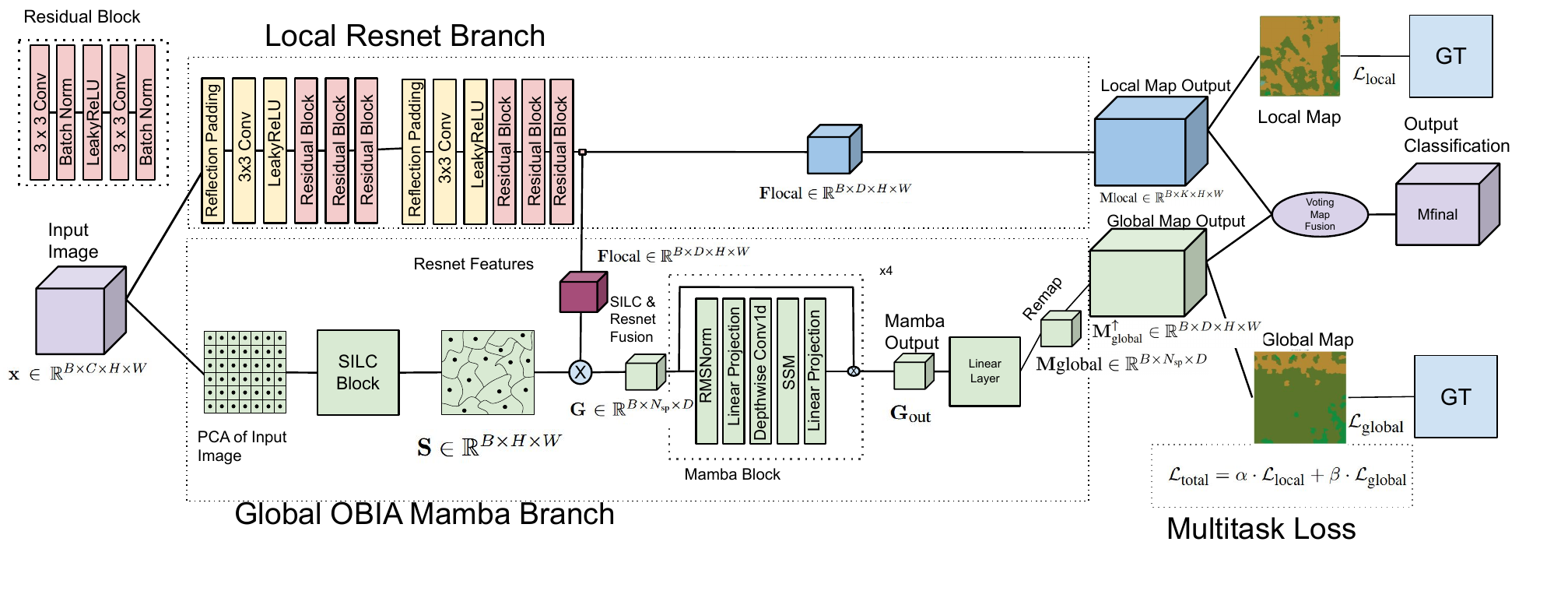} 
    \caption{The proposed OBIA Mamba is a GLocal dual branch architecture that features a local resnet branch and a global OBIA-Mamba. The OBIA-Mamba branch leverages superpixels, which reduces redundant computation by replacing pixel scanning with superpixel scanning. The GLocal architecture is joined together to produce a map that jointly models local spatial details and global contextual information. A multitask loss guides both the local and global branches to balance the local precision and global consistency.}
    \label{fig:overview}
\end{figure*}



This paper presents a novel Multitask Glocal OBIA-Mamba (MSOM) for enhanced Sentinel-2 classification with the following contributions.
\begin{itemize}
    \item First, an OBIA Mamba model (OBIA-Mamba) is designed to reduce redundant computation without compromising fine-grained details by using superpixels as Mamba tokens, seen in Figure \ref{scanning}. 
    \item Second, a global-local (GLocal) dual-branch CNN-mamba architecture is designed to jointly model local spatial detail and global contextual information.
    \item Third, a multitask optimization framework is designed to employ dual loss functions to balance local precision with global consistency.
\end{itemize}

The proposed approach is tested on Sentinel-2 imagery in Alberta, Canada, in comparison with several advanced classification approaches, and the results demonstrate that the proposed approach achieves higher classification accuracy and finer details that the other state-of-the-art methods.

The remainder of the paper is organized as follows. Section II illustrates the details of the proposed MSOM approach. Section III presents the experimental design and results. Section IV concludes this study.

\section{Methodology}

\label{methodology}
\subsection{Model Overview}
Figure \ref{fig:overview} shows that proposed model is a dual branch Global-Local (GLocal) architecture, with a local resnet branch and a global OBIA-Mamba branch, and a feature fusion module at the last stage to produce a final output. A multi-task loss is designed to supervise both the output from local branch and the output from the global branch. The input $Y$ is a sentinel-2 patch denoted by $\mathbf{x} \in \mathbb{R}^{B \times C \times H \times W}$, with the output map of size $H \times W \times K$, where  $K$ is the number of classes. 

\subsection{Global OBIA-Mamba Module}
The OBIA-Mamba module begins with object-based segmentation to change input images into  homogeneous regions. We extract the first 3 principal components (PC) of the Sentinel-2 image, which are used as input to the Simple Linear Iterative Clustering (SLIC) algorithm to generate superpixels. The resulting superpixels are represented by  $\mathbf{S} \in \mathbb{R}^{B \times H \times W}$, where each superpixel out of the $B$ superpixels contain pixels of similar spectral characteristics. 

The OBIA-Mamba module processes these superpixel regions to capture global contextual relationships through state space modeling. This object-based approach offers several advantages:  
(1) significant computational efficiency by reducing sequence length from $H \times W$ to $N_{\text{sp}}$, where typically $N_{\text{sp}} \ll H \times W$ (e.g., from 16,384 pixels to 500 superpixels for $128 \times 128$ images, a $32.8\times$ reduction);  
(2) built-in aggregation of local spatial context within each superpixel region; and  
(3) explicit modeling of relationships between perceptually coherent image objects rather than arbitrary pixel neighborhoods.

Given input features $\mathbf{F}_{\text{local}} \in \mathbb{R}^{B \times D \times H \times W}$ from the local branch and superpixel segmentation maps $\mathbf{S}$, the module first aggregates features within each superpixel region through spatial averaging:
\begin{equation}
\footnotesize
\mathbf{g}_i = \frac{1}{|\mathcal{P}_i|} \sum_{(h,w) \in \mathcal{P}_i} \mathbf{F}_{\text{local}}[h,w]
\end{equation}
where $\mathcal{P}_i = \{(h,w) \mid \mathbf{S}[h,w] = i\}$ denotes the set of pixel coordinates belonging to superpixel $i$, and $|\mathcal{P}_i|$ returns the number of pixels in this superpixel. This aggregation produces superpixel-level features $\mathbf{G} \in \mathbb{R}^{B \times N_{\text{sp}} \times D}$, with $N_{sp}$ being the number of superpixels. 

To handle variable numbers of superpixels across images while maintaining batch processing efficiency, features are zero-padded to a fixed superpixel numbers $N_{\text{sp}}^{\text{max}}$.

The core processing consists of four sequential Mamba blocks that progressively refine superpixel representations through long-range dependency modeling:
\begin{equation}
\footnotesize
\mathbf{G}_{\text{out}} = \text{MambaBlock}_4(
    \text{MambaBlock}_3(
        \text{MambaBlock}_2(
            \text{MambaBlock}_1(\mathbf{G})
        )
    )
)
\end{equation}
Each Mamba block employs a residual connection with RMS normalization and projects features to an expanded dimension $D' = 2D$ before applying depthwise convolution and selective state space modeling. The sequential processing enables each superpixel to aggregate contextual information from preceding superpixels in the sequence, building progressively richer representations that capture global image structure.

For segmentation output, global predictions are computed via linear projection:
\begin{equation}
\footnotesize
\mathbf{M}_{\text{global}} = \text{Linear}_{D \to K}(\mathbf{G}_{\text{out}}) \in \mathbb{R}^{B \times N_{\text{sp}} \times K}
\end{equation}
where $K$ is the number of semantic classes. These superpixel-level predictions are then remapped to dense pixel space using the original segmentation map $\mathbf{S}$:
\begin{equation}
\footnotesize
\mathbf{M}_{\text{global}}^{\uparrow}[b,c,h,w] = \mathbf{M}_{\text{global}}[b,\mathbf{S}[b,h,w],c]
\end{equation}
This remapping ensures spatial consistency within each superpixel region—all pixels belonging to the same superpixel share identical class predictions, maintaining the object-based nature of the representation while producing dense segmentation outputs.

\subsection{Local-Global CNN-Mamba Architecture}
The proposed architecture adopts a dual-branch design that combines local pixel-level processing with global superpixel-level reasoning. This hierarchical approach addresses a fundamental challenge in semantic segmentation: balancing fine-grained detail preservation with broader contextual understanding.

The local branch employs a ResNet-based feature extractor consisting of two cascaded ResNet modules. The two modules transform the input image $\mathbf{x} \in \mathbb{R}^{B \times C \times H \times W}$ into a feature maps $\mathbf{F}{\text{local}} \in \mathbb{R}^{B \times D \times H \times W}$ and an initial pixel-wise segmentation map $\mathbf{M}{\text{local}} \in \mathbb{R}^{B \times K \times H \times W}$, where $K$ denotes the number of semantic classes. This feature map $\mathbf{F}{\text{local}} \in \mathbb{R}^{B \times D \times H \times W}$ is also fed into the mamba processing block alongside the superpixels in the Global OBIA-Mamba branch. This local pathway captures low-level visual patterns including texture, edges, and fine spatial details that are critical for precise boundary delineation.

The global OBIA-Mamba branch runs alongside with the local branch and is as described in subsection B.

After both branches are complete, the final segmentation prediction employs an ensemble strategy through additive voting:
\begin{equation}
\footnotesize
\mathbf{M}{\text{final}} = \mathbf{M}{\text{local}} + \mathbf{M}{\text{global}}^{\uparrow}
\end{equation}
where $\mathbf{M}{\text{global}}^{\uparrow}$ denotes the global superpixel predictions remapped to pixel space. This voting mechanism allows the network to leverage both the precise localization capability of the local branch and the semantic coherence enforced by the global branch, resulting in classifications that are both spatially accurate and contextually consistent.

\subsection {Multitask Loss Learning}

The proposed architecture employs a dual loss learning strategy that jointly optimizes both local pixel-level predictions and global superpixel-level reasoning. This approach ensures that the network maintains fine-grained spatial accuracy while leveraging global contextual information.

The overall loss function combines local and global supervision through a weighted sum:

\begin{equation}
\footnotesize
\mathcal{L}_{\text{total}} = \alpha \cdot \mathcal{L}_{\text{local}} + \beta \cdot \mathcal{L}_{\text{global}}
\end{equation}

where $\alpha = 0.7$ and $\beta = 0.3$ are empirically determined weighting coefficients that prioritize pixel-level accuracy while maintaining global consistency.

The local loss $\mathcal{L}_{\text{local}}$ is computed between the initial pixel-wise segmentation map $\mathbf{M}_{\text{local}}$ and the ground truth masks using cross-entropy  loss. 
Global loss $\mathcal{L}_{\text{global}}$ supervises the superpixel-level predictions after processing through the Mamba blocks. 



\section{Results and Analysis}

\vspace{-0.8em} 


\subsection{Dataset}
\vspace{-0.5em}
For this letter, Sentinel-2 L2A imagery was acquired from July 2020 for the Canadian province of Alberta. To create a testing dataset, $4424$  $128$x$128$ patches are extracted from the full image. A 10\%/10\%/80\% train/val/test split is used. For ground truth, the 30 m NRCan Map resampled to 10 m is used. To overcome the noise and limitations in the 30 m map, the patches are only selected if the corresponding class dominates the image (i.e., takes up over 50\% of the pixels in the 128×128 patch). This allows the model to learn from the purest samples per class. The Sentinel-2 Scene Classification Layer (SCL) layer was also used to ensure no patches contained clouds or shadows.

\subsection{Implementation schema}

We compare the proposed method with various state-of-the-art approaches, i.e., HRNet , Swin Transformer, ResNet, ViT, LSTM, SSRN, ConvNeXt, and RNN. Overall accuracy (OA), averaged accuracy (AA), and the kappa coefficient are used for evaluating the results. For our method, we use a batch size of $32$, a learning rate of $0.001$,  epochs of $50$ and hidden dimensions of $64$. 

\begin{figure*}[!t]
    \centering
    \includegraphics[width=1.1\linewidth]{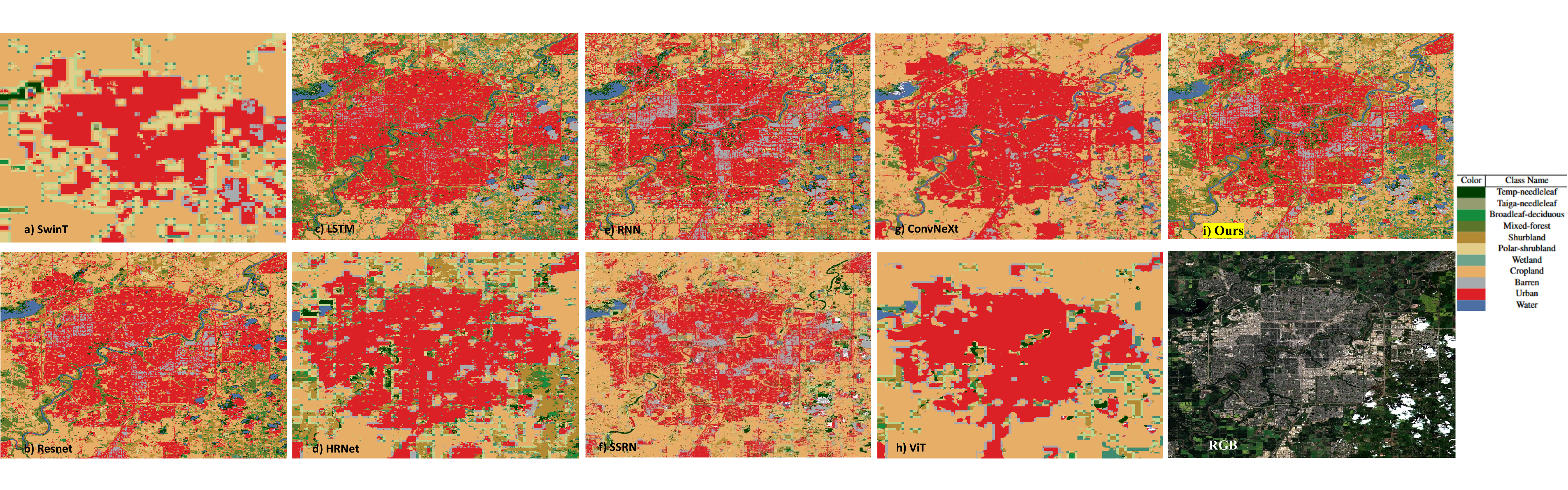}
    \caption{The maps generated by the various models on the city of Edmonton Alberta. It can be seen that our model provides the most detail compared to the RGB image, while simultaneously having the least amount of noise. It can be seen that our model depicts the entire river flowing through the city, which other models struggle to do. Our approach also demonstrates strong edge preservation, allowing prediction of small green patches throughout the city where natural classes can be observed in the RGB image.   }
    \label{fig:edmonton}
\end{figure*}

\begin{figure}
    \centering
    \includegraphics[width=1\linewidth]{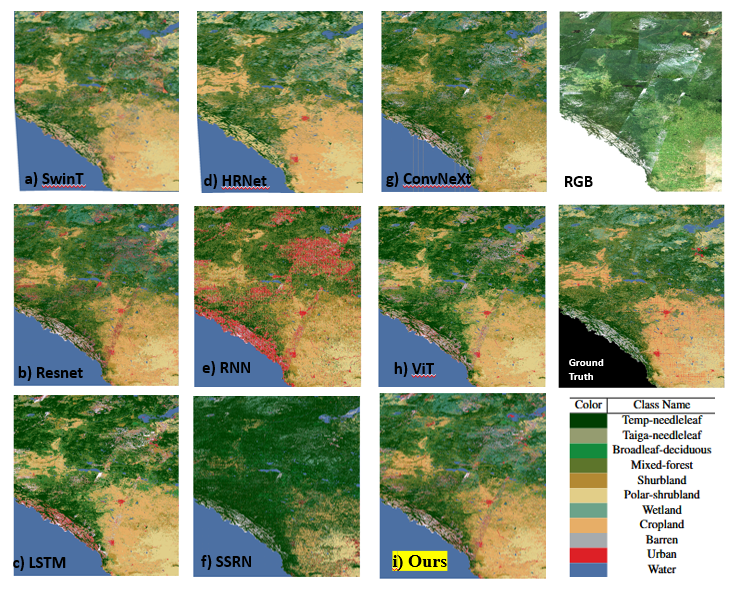}
    \caption{The province of Alberta generated by the various models. It can be seen that our model has the best spatial consistency with the ground truth. Our model does not suffer from over prediction on the urban class to the same extent as models such as RNN, LSTM and ResNet. The wetland class is also not underestimated in Northern Alberta in our approach unlike SSRN, HRNet and ViT. }
    \label{fig:fullmap}
\end{figure}

\subsection{Results}

\begin{table*}
\centering
\caption{Comparison of different methods on land cover classification. The best results are in bold with color shading.}
\scriptsize
\begin{tabular}{clc|ccccccccc}
\toprule
\multicolumn{3}{c|}{\textbf{Land Cover Class}} & \multicolumn{9}{c}{\textbf{Methods}} \\
\midrule
Color & Class Name & No. & HRNet & SwinT & ResNet & ViT & LSTM & SSRN & Convnext & RNN & \textbf{Ours} \\
\midrule
\cellcolor[RGB]{1, 62, 2} & Temp-needleleaf & 1 & 89.33 & 75.78 & 87.61 & 86.57 & 86.81 & 79.73 & \cellcolor[RGB]{251, 228, 213}\textbf{89.90} & 78.61 & 89.29 \\
\cellcolor[RGB]{149, 156, 112} & Taiga-needleleaf & 2 & 75.19 & 74.98 & \cellcolor[RGB]{251, 228, 213}\textbf{88.12} & 85.13 & 85.26 & 78.87 & 87.97 & 81.19 & 87.89 \\
\cellcolor[RGB]{20, 139, 61} & Broadleaf-dec. & 3 & \cellcolor[RGB]{251, 228, 213}\textbf{47.56} & 33.75 & 17.36 & 22.62 & 19.00 & 0.00 & 29.09 & 0.00 & 39.44 \\
\cellcolor[RGB]{93, 117, 43} & Mixed-forest & 4 & 72.75 & 81.47 & \cellcolor[RGB]{251, 228, 213}\textbf{87.22} & 84.47 & 85.18 & 86.99 & 87.04 & 66.29 & 83.05 \\
\cellcolor[RGB]{179, 137, 51} & Shrubland & 5 & 84.03 & 83.63 & 79.58 & \cellcolor[RGB]{251, 228, 213}\textbf{85.36} & 80.71 & 64.63 & 79.66 & 64.01 & 80.69 \\
\cellcolor[RGB]{226, 206, 136} & Grassland & 6 & 84.50 & 83.73 & 86.03 & 82.36 & 85.57 & 87.37 & 87.05 & \cellcolor[RGB]{251, 228, 213}\textbf{88.83} & 85.65 \\
\cellcolor[RGB]{200, 200, 200} & Polar-grassland & 7 & 0.00 & 0.00 & 0.00 & 0.00 & 0.00 & 0.00 & 0.00 & 0.00 & 0.00 \\
\cellcolor[RGB]{108, 163, 138} & Wetland & 8 & 79.46 & 81.88 & 85.69 & 83.12 & 86.63 & 75.65 & 83.87 & 81.31 & \cellcolor[RGB]{251, 228, 213}\textbf{89.83} \\
\cellcolor[RGB]{231, 174, 103} & Cropland & 9 & 89.56 & \cellcolor[RGB]{251, 228, 213}\textbf{90.14} & 79.72 & 83.12 & 67.97 & 85.33 & 90.29 & 57.11 & 87.96 \\
\cellcolor[RGB]{166, 171, 174} & Barren & 10 & 87.17 & 88.45 & 90.87 & 90.15 & 85.22 & 79.63 & \cellcolor[RGB]{251, 228, 213}\textbf{94.55} & 88.55 & 93.42 \\
\cellcolor[RGB]{221, 32, 38} & Urban & 11 & 54.29 & 86.00 & 67.18 & 90.68 & 72.84 & 56.07 & 73.16 & 48.47 & \cellcolor[RGB]{251, 228, 213}\textbf{91.23} \\
\cellcolor[RGB]{76, 112, 164} & Water & 12 & 95.23 & 94.84 & 96.59 & 95.59 & 96.10 & 96.59 & \cellcolor[RGB]{251, 228, 213}\textbf{97.34} & 95.27 & 96.52 \\
\cellcolor[RGB]{255, 255, 255} & Snow/Ice & 13 & 88.77 & \cellcolor[RGB]{251, 228, 213}\textbf{96.35} & 90.87 & 94.99 & 93.69 & 89.99 & 90.56 & 80.27 & 90.18 \\
\midrule
\multicolumn{3}{c|}{OA (\%)} & 82.95 & 82.16 & 81.87 & 83.54 & 80.61 & 76.13 & 84.50 & 71.76 & \cellcolor[RGB]{251, 228, 213}\textbf{85.88} \\
\multicolumn{3}{c|}{AA (\%)} & 75.22 & 74.70 & 73.60 & 75.70 & 72.71 & 67.77 & 76.19 & 63.84 & \cellcolor[RGB]{251, 228, 213}\textbf{78.08} \\
\multicolumn{3}{c|}{Kappa (\%)} & 81.24 & 80.41 & 80.02 & 81.89 & 78.65 & 73.69 & 82.92 & 68.88 & \cellcolor[RGB]{251, 228, 213}\textbf{84.47} \\
\bottomrule
\end{tabular}
\label{method_comparison}
\end{table*}

\begin{table}[h]
\centering
\scriptsize
\caption{Ablation Study of Different Map Outputs and Superpixels}
\label{tab:comparison}
\begin{tabular}{@{}lcccc@{}}
\toprule
\textbf{Metric} & \textbf{Local} & \textbf{Global} & \textbf{Voting} & \textbf{w/o Superpixels} \\
\midrule
OA     & 83.89 & 84.92 & \textbf{85.88} & \textcolor{gray}{75.62} \\
mIoU   & 66.11 & 67.21 & \textbf{68.77} & \textcolor{gray}{54.78} \\
Kappa  & 82.27 & 83.41 & \textbf{84.47} & \textcolor{gray}{73.20} \\
\bottomrule
\end{tabular}
\label{Maps}
\end{table}

\begin{table}[h]
\centering
\scriptsize
\caption{Ablation study on loss ratios (Local:Global).}
\label{tab:loss_ratios}
\begin{tabular}{@{}lccccccc@{}}
\toprule
\textbf{Metric} & \textbf{70:30} & \textbf{60:40} & \textbf{50:50} & \textbf{40:60} & \textbf{30:70} & \textbf{100:0} & \textbf{0:100} \\
\midrule
OA (\%)    & \textbf{85.88} & 85.87 & 85.86 & 85.81 & 85.84 & 60.63 & 84.27 \\
AA (\%)    & 78.08 & \textbf{78.35} & 77.90 & 78.23 & 78.01 & 56.37 & 77.07 \\
Kappa (\%) & \textbf{84.47} & 84.45 & 84.42 & 84.39 & 84.41 & 59.33 & 82.93 \\
\bottomrule
\end{tabular}
\label{loss}
\end{table}

Figure \ref{fig:fullmap} shows that the proposed architecture achieves a map that is more consistent with the classification map than the other methods, which tend to over-estimate the urban class or under-estimate the wetland class in the north.

Figure \ref{fig:edmonton} shows that the proposed approach is more consistent with the RGB image than the other methods. The proposed method performs better in preserving small details due to the strong edge-preservation capability of the super-pixel approach. ViT, SwinT and, HRNet tend to have a lack of detail due to the use of regular-patches as tokens and the use of down- and upsampling in their architectures. Resnet maintains local details well, but lacks the same global consistency as the proposed method. 

Table \ref{method_comparison} shows the numerical results achieved by different methods on the dataset. Our approach outperforms the other methods on all metrics. In particular, our approach achieves much better results on AA, indicating that the proposed approach outperforms the other approaches in terms of preserving and classifying the small classes. All the models failed to classify the polar-grassland class likely due to it's small training sample size and small presence in Alberta.

Table \ref{Maps} compares the local map, global map and voting map achieved by our approach. The voting map produces the highest accuracies in all metrics, indicating that the voting approach can better leverage the large-scale spatial context and local details than the local or global approach. The final column of the table showcases the results of our model without using superpixels and using the ordinary mamba pixel scanning. It can be seen that there is a 10\% accuracy decrease in using the ordinary pixel scanning instead of superpixel scanning.

Table \ref{loss} shows the impact of changing the $\alpha$ and $\beta$ ratios in the multi-task loss function. Using only a local or global loss significantly decreases the OA, AA, and Kappa compared to using a balanced ratio. A 70:30 ratio provides the highest overall accuracy and kappa, with a 60:40 ratio having a slightly higher AA. 




\section{Conclusion}
In this paper, we have presented a Multitask Glocal OBIA-Mamba (MSOM) approach to enhance LULC classification from Sentinel-2 imagery. We have the following contributions. First, an OBIA-Mamba approach has been designed to learn optimal token representations, which not only reduces the computational burden of the Mamba model by shortening sequence lengths but also preserves critical edge information and semantic coherence through OBIA principles. Second, a GLocal dual branch CNN-Mamba architecture is created, leading to a feature learning framework where CNNs are dedicated to learning local spatial context information and Mamba modules focus on efficient global dependency modeling, respectively. Last, a multi-task learning framework was designed based on dedicated loss functions which can efficiently optimize both local detail preservation and global contextual understanding for LULC classification. The proposed approach was tested on an Alberta Sentinel-2 dataset and compared to various other state-of-the-art approaches, demonstrating that our approach outperformed others in terms of both accuracy and computational efficiency while maintaining robust performance on challenging real-world scenarios with uncertain ground-truth boundaries.


\bibliographystyle{IEEEtran}
\bibliography{IEEEabrv,references}

\end{document}